\documentclass{article}

     \PassOptionsToPackage{numbers, compress}{natbib}

     \usepackage[preprint]{neurips_2020}

\usepackage[utf8]{inputenc} 
\usepackage[T1]{fontenc}    
\usepackage{hyperref}       
\usepackage{url}            
\usepackage{booktabs}       
\usepackage{amsfonts}       
\usepackage{nicefrac}       
\usepackage{microtype}      

\usepackage{amsmath}
\usepackage{bm}
\usepackage{graphicx}
\usepackage{color}

\title{Scalable Partial Explainability in Neural Networks via Flexible Activation Functions}

\author{%
  Schyler Chengyao Sun\\
  AI Group, DARTeC\\
  Cranfield University\\
  \texttt{Schyler.Sun@cranfield.ac.uk} \\
  \And
  Chen Li\thanks{Joint first author.} \\
  AI Group, DARTeC\\
  Cranfield University\\
  \texttt{C.Li.21@cranfield.ac.uk} \\  
  \And
  Zhuangkun Wei \\
  School of Engineering\\
  University of Warwick\\
  \texttt{Zhuangkun.Wei@warwick.ac.uk} \\
  \And
  Antonios Tsourdos \\
  SATM\\
  Cranfield University\\
  \texttt{A.Tsourdos@cranfield.ac.uk} \\
  \And
  Weisi Guo\thanks{Corresponding author.} \\
  AI Group, DARTeC, Cranfield University\\
  Alan Turing Institute\\
  \texttt{Weisi.Guo@cranfield.ac.uk} \\
  \texttt{wguo@turing.ac.uk} \\
}

\begin{document}

\maketitle

\begin{abstract}
Achieving transparency in black-box deep learning algorithms is still an open challenge. High dimensional features and decisions given by deep neural networks (NN) require new algorithms and methods to expose its mechanisms. Current state-of-the-art NN interpretation methods (e.g. Saliency maps, DeepLIFT, LIME, etc.) focus more on the direct relationship between NN outputs and inputs rather than the NN structure and operations itself. In current deep NN operations, there is uncertainty over the exact role played by neurons with fixed activation functions. In this paper, we achieve partially explainable learning model by symbolically explaining the role of activation functions (AF) under a scalable topology. This is carried out by modelling the AFs as adaptive Gaussian Processes (GP), which sit within a novel scalable NN topology, based on the Kolmogorov–Arnold Superposition Theorem (KST). In this scalable NN architecture, the AFs are generated by GP interpolation between control points and can thus be tuned during the back-propagation procedure via gradient descent. The control points act as the core enabler to both local and global adjustability of AF, where the GP interpolation constrains the intrinsic autocorrelation to avoid over-fitting. We show that there exists a trade-off between the NN’s expressive power and interpretation complexity, under linear KST topology scaling. To demonstrate this, we perform a case study on a binary classification dataset of banknote authentication. Our model converge at better precision rate than state-of-the-art SVM algorithms which indicates that we do not make performance sacrifices in our approach. Meanwhile, by quantitatively and qualitatively investigating the mapping relationship between inputs and output, our explainable model can provide interpretation over each of the one-dimensional attributes. These early results suggest that our model has the potential to act as the final interpretation layer for deep neural networks.

\end{abstract}

\section{Introduction}

Improving our trust in deep neural networks (DNN) can be achieved by developing the statistical and explainable foundations. Trust in deep learning is important, because of the increasing widespread use of commercial DNN based artificial intelligence (AI), especially in the areas that engage with human life such as autonomous vehicles and bank transactions \cite{whitepaper}. There are a multitude of potential risks (e.g. discrimination, adversarial attacks, over-fitting) \cite{8611298} and explainable AI \cite{MeijerG} has the potential to both offer insight during the operations and in a post-hoc manner. 

\subsection{Motivation and Related Work}

At the heart of our need to add explainability / interpretability to DNNs is the need to build trust in a quantifiable way. Traditional mathematical model-based algorithms have reasonably high clarity in how a model and the input data leads to output decisions. Bayesian methods can quantify the uncertainty both in the forward and inverse problem. Whilst DNNs have been shown to be able to accelerate the solution discovery of many iterative optimisation problems \cite{XAI6GUO}, they remain opaque and doesn't tell us the impact of input data and bias on decisions, the reasoning for decisions, and how the DNN logic can reverse teach human experts. 

Beyond these technical requirements, the legal framework for AI is still in its infancy, and there are several explicit requirements for XAI in different regions, such as EU GDPR (see Recital 71) requires machine learning algorithms to be able to explain their decisions. The key is that rightly or wrongly, humans can attempt to explain if prompted to, and we need DNNs to have that equal capability in order to ensure trust and a legal pathway towards improving safety and reliability.

The most common way to achieving explainability for NN is to evaluate the impact of each input on the output, e.g. Saliency maps\cite{saliency}, DeepLIFT\cite{DeepLIFT}, LIME\cite{LIME} can obtain the approximate solution to provide aforementioned type of explanation by reverse analysis for instances. However, these methods are more focusing on the direct relation between inputs and output rather than the NN structure and inner operations. Authors of \cite{MeijerG} propose to demystify black-box models with symbolic meta-models which leads a pathway to split and explicit the inner operations of NN and inspired us to improve transparency in NNs from an activation function and topology perspective.

\paragraph{Flexible Activation Function}
\label{problems}
Conventional NN typically have a fixed, bounded continuous non-linear activation function (AF) at each neuron, which is the key to the overall nonlinear behavior of NN. However, with the fixed AF, the expressive power of each layer is capped \cite{NNsCap0,NNexpress,NNsCap1}, thus the neural network can only become deeper in order to fit complex high dimensional nonlinear data. Therefore, there is a strand of works \cite{AFlinear1,AFPoly,AFCubicSpline0,AFCubicSpline,BSpline,AFGP,AFtunable,AFtunable2,AFCNN} tried to train the NN by tuning AF for purpose of enhancing the expressive power of nodes so that to reduce the topological complexity. The core idea of flexible AF is using control parameters for the curve shaping, which can be considered as interpolation between control points, while these parameters being optimized during the back-propagation process. 

Interpolation methods have applied different AFs: (i) Piecewise linear interpolation \cite{AFlinear1} has the best flexibility, but easily leads to overfitting. Thus, the penalty function is crucial and sensitive to the model; (ii) Polynomial interpolation \cite{AFPoly} avoids the overfitting to some extent due to the constraint of the function, but the local flexibility is sacrificed; (iii) Spline interpolation \cite{AFCubicSpline0,AFCubicSpline,BSpline,BSP} has both locally and globally flexibility while avoids the overfitting, but the splines are hard to express in terms of symbolic functions, which is the gateway to all explanations \cite{MeijerG}; (iv) Gaussian Processes (GP) interpolation \cite{AFGP,GPN} is a non-parametric model that can give the symbolic expression of AFs, while offer an expressive power vs. explainability trade-off in training set from the kernel space. Due to the kernel function constrains, GP AFs can effectively prevent overfitting as well as give a symbolic function that conforms to intrinsic autocorrelation which is better than others for explanation purpose. Therefore, in our proposed NN, we make each AF be generated by GP interpolation between control points within it, which can thus be tuned during the back-propagation procedure via gradient descent. By visualizing the activation, we can decrease difficulty in explaining the role of each node and layer in NN.  

\paragraph{NN Topology}
In common NN, topology is chosen from a set of known models (AlexNet, VGGNet, GoogleNet, etc) or customized with few limitations \cite{top1,top2,top3}. Moreover, from the field of evolutionary computing, search algorithms for neural network topologies are proposed in \citep{topology1,topology2,topology3,topology4,topology5}.
However, dynamic and complex topologies are not suitable for our main purpose -- Explainability. In this paper, we propose lay our explainability scheme on a fixed topology mode. \cite{Kolmogorov,KST1,KST2} shows that Kolmogorov–Arnold Superposition Theorem (KST) can offer an approximation to any continuous function in high dimensional space using a finite composition of (a) univariate continuous functions and (b) addition operation. Therefore, based on KST, we establish a scalable NN topology as the foundation of our explainability since both (a) and (b) are the basic elements of all operations which are easier for understanding.

\subsection{Novelty and Contribution}
In this paper, we approach the problem of NN explainability framework by introducing flexible activation functions into Kolmogorov–Arnold Superposition Theorem (KST) based topology NN for interpreting its inner workings in terms of symbolic and visualized functions for each neuron. Under our proposed model, we achieve global transparency to the NN and show the trade-off between NN expressive power vs. explainability. The remainder of this paper is organised as follows. In Section 2, we build a system model step by step. In Section 3, we apply the model to the a binary classification dataset of banknote authentication for case study and conduct the explainability analysis of it. Section 4 concludes this paper and proposes the ideas for future work.

\section{System Model}
\label{Model}
\subsection{Neural Network Topology}
\paragraph{Kolmogorov–Arnold Superposition Theorem}
Hilbert’s 13th problem solved by Kolmogorov–Arnold superposition theorem (KST) \cite{Kolmogorov}, neural networks (NN) can prove to be universal approximators for every continuous function mapping \cite{SigProve,universal}. There are many generalizations and refinements of KST. We state one of these, which uses the primary work in \cite{Kolmogorov}. To be specific, for any $D \in \mathbb{N}$, there exist $R \leq 2D$ and continuous functions $\phi_{rd}(\lambda_d):\mathbb{I}\xrightarrow{}\mathbb{R}$ for $d=1,2,...,D$ and $r=0,1,...,R$, such that: for every arbitrary multivariate continuous function $f(\bm{\lambda}):\mathbb{I}^D\xrightarrow{}\mathbb{R}$, where $\bm{\lambda}=[\lambda_1,\lambda_2,...,\lambda_d,...,\lambda_D]^\mathrm{T}$ there exist continuous functions $\Phi_r:\mathbb{R}\xrightarrow{}\mathbb{R}$ for $r=0,1,...,R$, such that we may define:
\begin{equation}
\label{Superposition}
    F(\bm{\lambda})=\sum^R_{r=0}\Phi_r \left(\sum^D_{d=1}\phi_{rd}(\lambda_d)\right)
\end{equation}
as an approximate realization of function $f(\bm{\lambda})$; that is, given any $\epsilon>0$, $|F(\bm{\lambda})-f(\bm{\lambda})|<\epsilon$ for each $\bm{\lambda}\in\mathbb{I}^D$, which means functions of the form $F(\bm{\lambda})$ are dense in $C(\mathbb{I}^D)$. Kolmogorov also showed that the inner functions $\phi_{rd}$ are independent with the outer functions $\Phi_{r}$ may have latent dependence of the target function $f$.

\paragraph{KST-Based NN Topology}
We construct our initial NN topology model according to the state of KST.
Each neural node contains an adjustable activation function which maps the sum of inputs into its output. We set the $\underline{R}=0,1,2,...$ in (\ref{Superposition}), the \emph{\underline{repetition level}}, as the only parameter of our proposed topology, which control scaling as well as the trade-off between potential approximation ability and width of the NNs model while the depth of NN is fixed. Meanwhile, we define each repetition in topology as a \emph{\underline{unit}}, and the final estimated output $\hat{\gamma}$ is the sum of $R+1$ units. This enable us to alleviate the complexity of explanation into a fixed mode. Schematic diagrams in Fig.\ref{top} demonstrate two topology examples and indicate that the KST topology is the key to the trade-off between model expressive power and explainability while flexible activation function is applied to enhance the expressive power of each node -- as we will show in Section 2.2.

\begin{figure}
  \centering
  \includegraphics[width=1.0\linewidth]{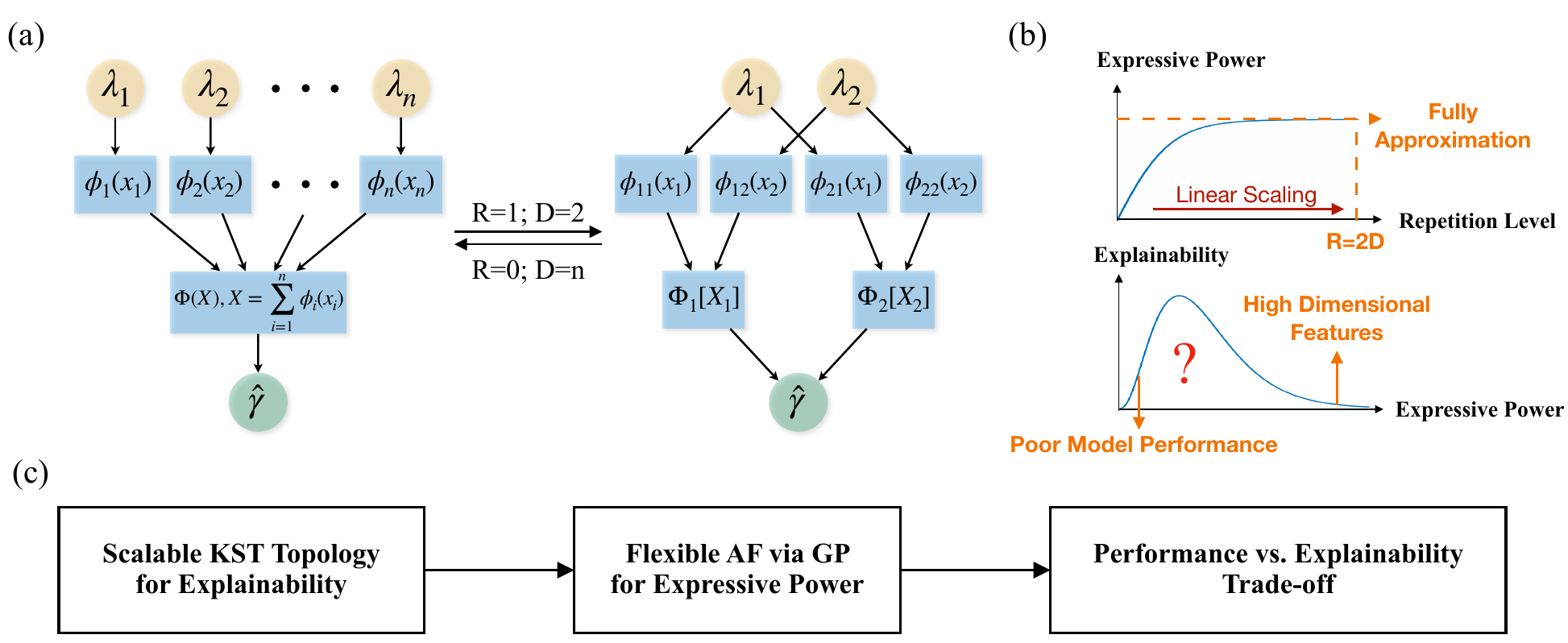}
  \caption{(a) The KST based neural network topology. $\Phi$ and $\phi$ are the flexible GP AFs. Left gives an example for one unit with $n$ dimensions inputs while right is two units topology for two dimensions inputs; (b) Qualitative relationship within expressive power, scaling and explainability in our model; (c) Our working flow for scalable partial explainability in neural networks via flexible AFs.}
  \label{top}
\end{figure}

\subsection{Activation Function}
Different from the conventional NN using fixed AF while training the additional weight and bias of each input for the neural node, we use unmodified input for each proposed neural node where an adjustable AF is within.
In other words, we embed the weight and bias of each input into the AF it is designated, and at the same time, tuning the shape of AF.

In order tackle the aforementioned problem in \ref{problems} (i.e. function discontinuity, local tuning difficulty, parameter dependency and model over-fitting), we proposed to apply the noise-contained Gaussian Processes (GP) to fit the AF with control points. Specifically, each AF is generated by a GP regression of the control points while we tolerate the existence of noise on control points. 

In this case, we can achieve the following objectives for the AF: (1) Ensure of the intrinsic autocorrelation within the function for smoothness and explainability; (2) Gain both local and global function adjustability; (3) Avoid over-fitting.


\paragraph{Priori Gaussian Processes}
Consider each activation function is assumed to follow a latent GP plus noise which can be expressed as:
\begin{equation}
\label{latentGP}
	\phi(x)=\mathrm{GP}(x)+\epsilon
\end{equation}

where $\mathrm{GP}(x)$ is the random variable (RV) which follows a distribution given by GP, and $\epsilon$ is the additive Gaussian noise with zero mean and variance $\sigma^2$. From the continuous AF domain, finite number of control points are taken as $ \bm{x}=[x_1,x_2,...,x_n]^\mathrm{T}$, with $\bm{y}=\phi(\bm{x})= [y_1,y_,...,y_n]^\mathrm{T}$, can be assumed to follow the multivariate Gaussian as
\begin{equation}
\label{multiG}
    \bm{\mathrm{GP}(x)} \sim \mathcal{N} (\bm{\mu(x)},\bm{K(x,x)})
\end{equation}
where $\bm{\mu(t)}$ is the mean function and $\bm{K(x,x)}$ is the covariance matrix given by:
\begin{equation}
\bm{K(x,x)}= \left[
    \begin{matrix}
    k(x_1,x_1)     	& k(x_1,x_2)       & \cdots 	& k(x_1,x_n)      \\
    k(x_2,x_1)      & k(x_2,x_2)      & \cdots 		& k(x_2,x_n)      \\
    \vdots 			& \vdots 		  & \ddots 	    & \vdots \\
    k(x_n,x_1)      & k(x_n,x_2) 	& \cdots 		& k(x_n,x_n)  \\
    \end{matrix} \right]
\end{equation}
where $k(x_i,x_j)$ is the covariance between RVs $\mathrm{GP}(x_i)$ and $\mathrm{GP}(x_j)$ represented by the kernel function.
Thus, according to \ref{latentGP} and \ref{multiG}, the priori GP probability model
can be expressed as:
\begin{equation}
    \bm{y} \sim \mathcal{N} (\bm{\mu(x)},\bm{C(x,x)}).
\end{equation}
where $\bm{C(x,x)}=\bm{K(x,x)}+\sigma^2_n \bm{I_n}$.

\paragraph{Kernel Function}
In GP, the covariance between every two RVs is quantified by the kernel function which interprets the potential correlation between RVs. Several appropriate kernels can be selected for different priori knowledge, e.g. smooth curve is obtained by fitting with radial-basis function (RBF) kernel and exp-sine-squared kernel is designed for periodic patterns.
In our proposed activation function, we aim for two goals, i.e. (1) Global autocorrelation with local adjustability; (2) Continuity and smoothness, so that rational quadratic (RQ) kernel would be a default choice in our experiments:
\begin{equation}
		k(x_i,x_j)=\sigma^2(1+ \frac{({x_i}-{x_j})^2}{2\alpha l^2})^{-\alpha}
		\hspace{1cm}
		i,j=1,2,..,n
\end{equation} 
where $\sigma^2$ determines the variance magnitude, $l$ is length-scale parameter and $\alpha$ is scale mixture parameter. RQ kernel can be considered as a combination of infinite sum of RBF kernels with various length-scales hence to vary smoothly across multiple length-scales which makes it more flexible than the RBF kernel to fit local adjustment of control points and to alleviate gradient vanishing problem while maintain the noise resistance property of GP\cite{GPBook}. Further more, various common and customized kernels alternatives may also be applicable in some specific scenario with priori knowledge embedded \cite{kernel1,kernel2,kernel3}.

\paragraph{Posterior Gaussian Processes}

Kernel function selected, the hyper-parameters $\bm{\theta}$ of the kernel can be tuned by maximizing the corresponding log marginal likelihood function which is equivalent to minimizing the cost function:
\begin{equation}
	\arg\min_{\bm{\theta}}l(\bm{\theta})=\bm{y}^\mathrm{T} \bm{C}^{-1} \bm{y}+ \log \left| \bm{C} \right|
\end{equation} 
The conventional quasi-Newton and gradient descent methods can be used in this optimization problem while \cite{TrafficGP} also offers an algorithm for reducing the computational complexity of hyper-parameter learning from $\mathcal{O}(n^3)$ to $\mathcal{O}(n^2)$ without performance loss.

In the kernel hyper-parameters tuning process, due to the allowance of noise $\epsilon$ in $\phi(\bm{x})$, the AFs effectively avoid over-fitting, since the GP will follow the covariance constraints given by kernel and result in an curve which is not guaranteed to pass through all the control points. Therefore, the GP can give penalty to a potential over-fitted control point by considering it as noise, so that to keep itself still being an autocorrelation function.

Kernel hyper-parameters tuned and optimized, GP can give the posterior Gaussian distribution for every RV $y_*=\phi(x_*)$ within the AF domain with mean and variance as \cite{GP}:
\begin{equation}
\label{posterior}
     \overline{\phi(x_*)}=\bm{\mu}+\bm{k}^\mathrm{T}(\bm{x},x_*)\bm{C}^{-1}(\bm{y}-\bm{\mu}) \hspace{0.25cm}\mathrm{with}\hspace{0.25cm}
     V[\phi(x_*)]=k(x_*,x_*)-\bm{k}^\mathrm{T}(\bm{x},x_*)\bm{C}^{-1}\bm{k}(\bm{x},x_*)
\end{equation}
and here we can assume that $\bm{\mu}=\bm{0}$ \cite{GPCam}. Same as Bayesian deep learning, the GP AFs give not only the mean values but also the variance which lead the pathway to quantify the epistemic and aleatoric uncertainties of the model \cite{BayesianDL}. In this paper, we only use the mean for activation value and discuss the explainability.

\subsection{Back-Propagation for Control Points Tuning}
Feed-forward NN structure established, we can use back-propagation algorithm for control points' coordinate tuning so that to adjust the AF. Let $L(\bm{\Theta})$ denote the loss for a batch of instances, where $\bm{\Theta}=\{x_1,x_2,...y_1,y_2,...\}$.
In each epoch for a batch of training instances, we can perform an update on control points' coordinate with a learning rate $\eta$ as:
\begin{equation}
\left[
    \begin{matrix}
    x_i^{t+1}\\
    y_i^{t+1}\\
    \end{matrix} 
\right]
\leftarrow
\left[
    \begin{matrix}
    x_i^{t}\\
    y_i^{t}\\
    \end{matrix} 
\right]
-\eta
\left[
    \begin{matrix}
    \frac{\partial L(\bm{\Theta})}{\partial x_i}\\
    \frac{\partial L(\bm{\Theta})}{\partial y_i}\\
    \end{matrix} 
\right]
\end{equation}

Similar to conventional NN, the chain rule still works for loss gradient descent whatever loss function (e.g. cross entropy, MSE loss, etc.) is chosen while the key is to obtain the function slope $\kappa(x_*)=\frac{\partial \overline{\phi_*}}{\partial x_*}$ at each point $x_*$ and the gradient $\bm{\Delta_i}=[\frac{\partial \overline{\phi_*} }{\partial x_i},\frac{\partial \overline{\phi_*} }{\partial y_i}]$ for each control point, where $i=1,2,...,n$. Due to the complex optimization processes in GP regression, for calculation reduction, we use finite difference method in (\ref{posterior}) for $\kappa(x_*)$ and $\bm{\Delta_i}$ approximate solutions. We also perform some experiments in which $\kappa(x_*)$ and $\bm{\Delta_i}$ are given precisely, however the accuracy improvement is ambiguous.

\subsection{Symbolic Model and Visualized Explainability}
Using all trained AFs, the symbolic representation of the model (\ref{symbolic}) can be directly derived from (\ref{Superposition}) and (\ref{posterior}) which gives the transparency to the full mapping relationship. Furthermore, with figures of the AFs, we can backtrack how each instance inputs map to the output visually (e.g. Fig.2a). i.e. At the last layer, contribution from each unit can be evaluated, so that the reason why model perform well/bad in each instance can be attributed to some specific units. In the same way, we can discover which input features are decisive at the first layer. Moreover, we can easily reverse this model in order to explore the data group which lead to our concerned output. Therefore, the noise attributes and potential feature discrimination might be dug out. In the next section, we present a case study to illustrate how our proposed AFs offer the transparency and partial explainability for the neural network.

\section{Case Study and Result Discussion}
In this section, we perform a case study on a 1372 instances, four-attributes binary classification dataset for banknote authentication \footnote{Data Source: https://archive.ics.uci.edu/ml/datasets/banknote+authentication} with attribute information: ($\lambda_1$) variance of wavelet transformed image; ($\lambda_2$) skewness of wavelet transformed image; ($\lambda_3$) kurtosis of wavelet transformed image; ($\lambda_4$) entropy of image. Each attribute is standardized into $[-1,1]$ interval for clearer illustrating. Furthermore, we add an attribute with artificial noise ($\lambda_5$), which uniformly distribute over $[-1,1]$, in order to evaluate the model’s ability to deal with noise. We set two criteria to evaluate the model performance, namely the classification accuracy and the numerical loss. We use one for genuine and zero for forged to represent the authenticity of the banknote. We use the nearest neighbor principle for classification while calculate the numerical loss for back-propagation. The model and system setup of our experiments implementation are given below in Table I.

\begin{table}[h]
\renewcommand{\arraystretch}{1.2}
\centering
\caption{System Setup}\vspace{-1mm}
\begin{tabular}{c|c}
\hline
\bfseries Parameter             &   \bfseries Value \\
\hline\hline
Number of Units                       & 2   \\
Number of Control Points              & 6 for Each Neuron   \\
Control Points Initialization         & Random   \\
GP Interpolation Kernel                 & Rational Quadratic\\
Learning Rate                       & Layer 1: 1e-1; Layer 2: 1e-3 \\
Division                        & Random Division; Training: Validation (Test)=7:3 \\
Back-propagation Epochs                    &  1000 \\
\hline
Processor                                  & 2.3 GHz 8-Core Intel Core i9\\
Memory                                   & 16 GB 2400 MHz DDR4\\
\hline
\end{tabular}
\vspace{-2mm}
\label{Tab1}
\end{table}

\begin{figure}
  \centering
  \includegraphics[width=1\linewidth]{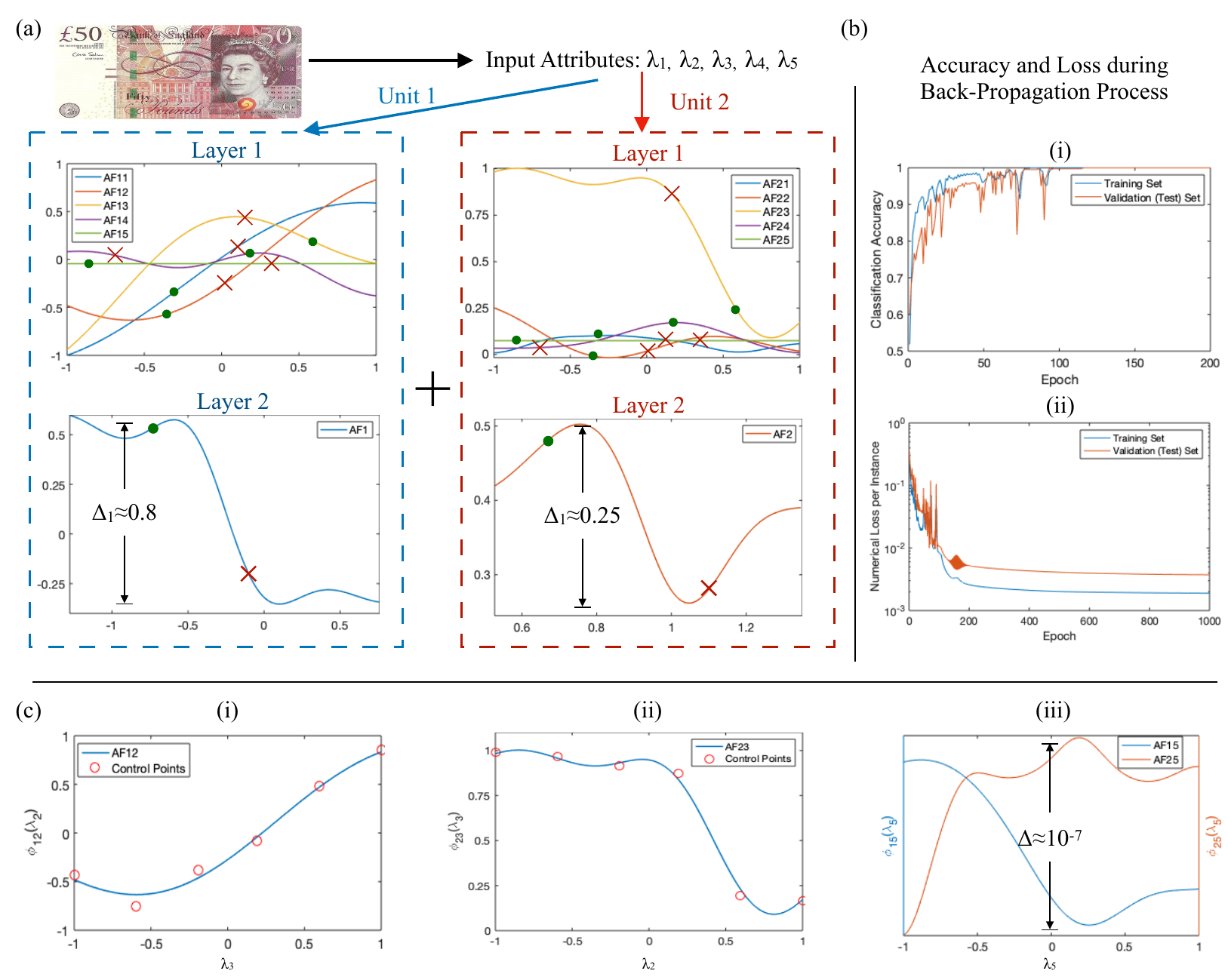}
  \caption{\textbf{Visual Explanation:} (a) The AFs in each neuron after training process are demonstrated. The green points and red crosses marked on the AFs show how two typical representative inputs from two classes data map to the output; (b) The model performance are evaluated with both the classification accuracy and the numerical loss during the back-propagation process; (c) Three noteworthy AFs are displayed separately for analysis.}
  \label{top}
\end{figure}

\paragraph{Results Overview}
Symbolic Explainability - After 428s running time, we obtain the symbolic trained model in the form of:
\begin{equation}
\label{symbolic}
    F(\lambda_1,\lambda_2,\lambda_3,\lambda_4,\lambda_5)=\sum^1_{r=0}AF_r \left(\sum^5_{d=1}AF_{rd}(\lambda_d)\right)
\end{equation}

Fig.2(a) visualizes the trained NN networks with every AF while we backtrack two typical representative data from two classes in the AFs with green points for the genuine banknote and red crosses for the forged one. Fig.2(b) shows the classification accuracy achieve 100\% for both training and validation set after around 120 epochs, which is higher than the classical SVM (99.2\%) on this dataset while the numerical loss is converged within $[10^{-3},10^{-2}]$ for each instance and no overfitting has occurred. Fig.2(c) focuses on three noteworthy AFs.

\paragraph{Model Interpretation}
In our model, the output result is the sum of the outputs from the two units. Consider that each unit represents a feature of the data which act as an additive to the model result, thus we separately analyze how these units effect the model result. At layer 2, the AF1 and AF2 (Fig.2a) give different value ranges with $\Delta_1\approx 0.8 > \Delta_2 \approx 0.25$ which indicates that the unit 1 has a more decisive impact than unit 2 on the model result, meanwhile it is worth mentioning that AF1 demonstrate a distinct trend towards binary classification, which such is the case in the previous description of KST -- the outer functions may have latent dependence of the target function. Firstly, focus on unit 1, it can be qualitatively observed that the $\lambda_1, \lambda_2$ and $\lambda_3$ are the main attributes which effect most on the unit 1 output. Fig.2c(i) gives an example for AF12 with its control points which start with a small decrease followed by an increase. Analyze it together with the layer 2, we can conclude that the lower $\lambda_2$ make the banknote more easily to be classified as genuine. Turn to unit 2, the $\lambda_3$ almost dominate on the output of unit 2.  With a rapid change over $[0,0.5]$ interval in AF23 (Fig.2c(ii)), $\lambda_3$ attribute has the ability to distinguish the two classes of data in unit 2. In both units, the impact of attribute $\lambda_4$ is not notably on the classification result of our model due to relatively narrow range of AF14 and AF24. Meanwhile, Fig.2c(iii) shows that our model has the robustness to noise attribute $\lambda_5$ -- AF15 and AF25 give extremely little contribution to the model result other than over-fit the data. 

In our model, the symbolic expressions can help data scientists with better understanding of the global mapping relationship within the NN. Furthermore, by analyzing the trained model reversely, scientists can anticipate potential risks in advance about how banknote would be forged to pass the detector under this case. 
Meanwhile, for AI users, the one-dimensional visual functions flow offer the transparency and partial explainability on how model result come from each attribute input, which can enhance the trust to the model.

\paragraph{Scalability, Expressive Power and Explainability}
Our model is linear scalable with the repetition level in topology and the number of control points in each neuron. The number of control points determines how much details can the function has while the units number determines the maximum number of potential features that can be extracted. Fig.3 shows that more units and more control points lead the model converging at lower numerical loss which means stronger model expressive power. However, the difficulty of model explaining for AI users is increasing at the meantime -- The AI users are not sensitive to small variations in the function (Fig.3a) while confuse the role of multiple units (Fig.3b). The ascent of model accuracy and model explaining difficulty are subject to different scales -- $log$ scale and linear (or even exponential) scale. Thus, we need to find a balance between these two or establish a trade-off based on different scenario requirements.

\begin{figure}
  \centering
  \includegraphics[width=1\linewidth]{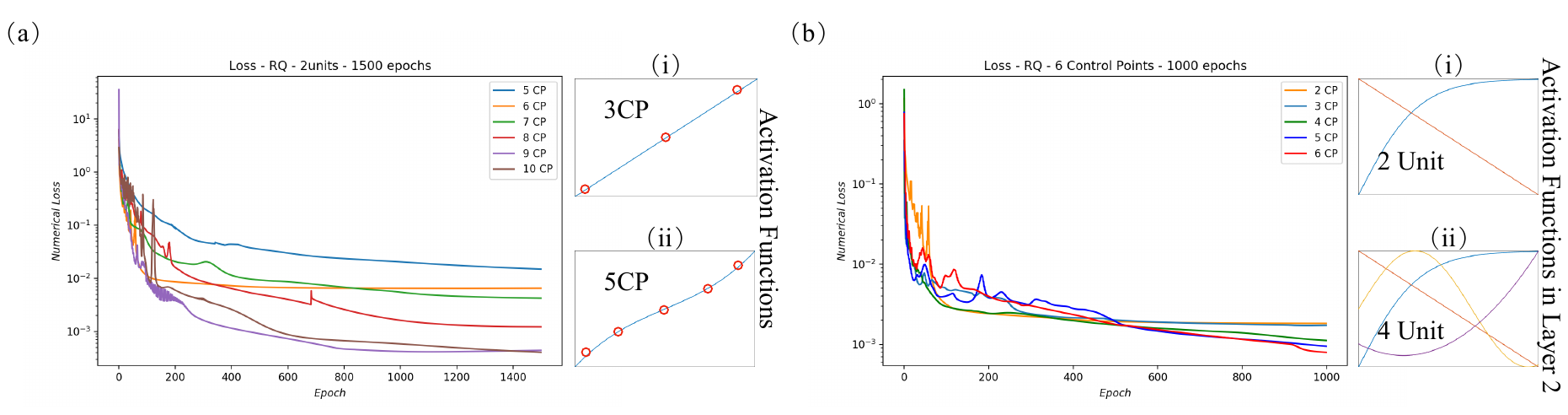}
  \caption{ More control points in each neural (a) and more units in topology (b) can lead to better model performance, however as such, the difficulty of model explaining for AI users will increase at the meantime.}
  \label{CP}
\end{figure}


\section{Conclusion and Future Work}

In this paper, we found that by modeling neural networks as a combination of linear scalable architectures via the Kolmogorov–Arnold Superposition Theorem (KST) and generalised Gaussian Process activation functions, we are able to demonstrate a new degree of interpretability. In particular, we have demonstrated a trade-off between expressive power and explainability of the NN through controlling the KST repetition level. This work expands on current state-of-the-art interpretation methods (e.g. Saliency maps, DeepLIFT, LIME, etc.) by focusing more on the role played by activation functions and the NN structure itself. 

To demonstrate applicability, we perform a case study on a binary classification dataset of banknote authentication. Our model converge at better precision rate than state-of-the-art SVM algorithms which indicates that we do not make performance sacrifices in our approach. Meanwhile, by quantitatively and qualitatively investigating the mapping relationship between inputs and output, our explainable model can provide interpretation over each of the one-dimensional attributes. These early results suggest that our model has the potential to act as the final interpretation layer for deep neural networks.

Our future work will focus on (i) fitting orthogonal features to different unit in order to compress our network by reducing the repetition level while keeping approximately the same model accuracy \cite{orthogonal,orthogonal2}; (2) establishing a pattern library of common features using activation functions for easier identifying the feature represented by a unit.

\bibliographystyle{unsrt} 
\bibliography{main}

\end{document}